\pdfoutput=1

\documentclass[11pt]{article}

\usepackage[final]{acl}

\usepackage{times}
\usepackage{latexsym}

\usepackage[T1]{fontenc}

\usepackage[utf8]{inputenc}

\usepackage{microtype}

\usepackage{inconsolata}

\usepackage{graphicx}
\usepackage{makecell}
\usepackage{etoolbox}
\AtBeginEnvironment{table}{\small}

\title{\textsc{NAVCON}: A Cognitively Inspired and Linguistically Grounded Corpus for Vision and Language Navigation}

\makeatletter
\def\blfootnote{\xdef\@thefnmark{}\@footnotetext}
\makeatother

\author{
 \textbf{Karan Wanchoo\textsuperscript{1}},
 \textbf{Xiaoye Zuo\textsuperscript{1}},
 \textbf{Hannah Gonzalez\textsuperscript{1}},
 \textbf{Soham Dan\textsuperscript{1}},
\\
 \textbf{Georgios Georgakis\textsuperscript{1*}},
 \textbf{Dan Roth\textsuperscript{1}},
 \textbf{Kostas Daniilidis\textsuperscript{1}},
 \textbf{Eleni Miltsakaki\textsuperscript{1}},
\\
 \textsuperscript{1}University of Pennsylvania,
\\
 \small{
   \textbf{Correspondence:} \href{mailto:kwanchoo@seas.upenn.edu}{kwanchoo@seas.upenn.edu}
 }
}

\begin{document}
\maketitle
\begin{abstract}
We present \textsc{NAVCON}, a large-scale annotated Vision-Language Navigation (VLN) corpus built on top of two popular datasets (R2R and RxR). 
The paper introduces four core, cognitively motivated and linguistically grounded, navigation concepts and an algorithm for generating large-scale silver annotations of naturally occurring linguistic realizations of these concepts in navigation instructions. We pair the annotated instructions with video clips of an agent acting on these instructions. \textsc{NAVCON} contains $236,316$ concept annotations for approximately $30,0000$ instructions and $2.7$ million aligned images (from approximately $19,000$ instructions) showing what the agent sees when executing an instruction. To our knowledge, this is the first comprehensive resource of navigation concepts.  We evaluated the quality of the silver annotations by conducting human evaluation studies on \textsc{NAVCON} samples.  As further validation of the quality and usefulness of the resource, we trained a model for detecting navigation concepts and their linguistic realizations in unseen instructions. Additionally, we show that few-shot learning with GPT-4o performs well on this task using large-scale silver annotations of \textsc{NAVCON}.
\end{abstract}

\section{Introduction}

\blfootnote{* Currently employed at the Jet Propulsion Lab, California Institute of Technology. This work was done as an outside activity and not in the author's capacity as an employee of the Jet Propulsion Laboratory, California Institute of Technology. }

The main goal of Vision-Language Navigation (VLN) tasks  is to enable robots or other agents to navigate in a variety of spaces following language instructions. Given the resource limitations inherent in robotic navigation, as well as the unpredictability of the navigation spaces,  successful VLN models must be (a) resource efficient in real time processing of  linguistic and visual inputs and (b) capable of applying their learning to previously unseen linguistic instructions in new physical environments.  
In the interest of understanding the non-negligible gap between machines’ performance and human benchmarks there is increasing motivation to move from \textsc{black box} end-to-end models to models that are grounded, that is, they are capable of demonstrating an understanding of the concepts they learn and the decisions they make.  
Currently, models such as CLIP-ViL~\cite{shen2021much} have shown unbalanced attention on text and vision~\cite{zhu2103diagnosing}, thus compromising the reliability of aligning text tokens to visual inputs.
To enable the training of more transparent models with regard to this cross-modal alignment, we need datasets with higher-level annotations of linguistically grounded navigation concepts and their associations to navigation action in the physical world. Creating sophisticated datasets with reliable annotations at the scale required for effective training is a well-known challenge that sometimes requires unrealistic human effort. 
In this paper, we address this challenge by introducing \textsc{NAVCON}\footnote{\footnotesize anonymous.4open.science/r/NAVCON-E068/}, a large scale VLN corpus annotated with high level navigation concepts, requiring minimal human annotation effort. We identify four core navigation concepts, inspired by cognitive functions and observations from VLN datasets, essential for conveying navigation instructions. These concepts, considered higher-order linguistic predicates, are grounded in diverse linguistic realizations from over $30,000$ instructions across two popular community-standard VLN datasets. We then ground these linguistic expressions with video clips of agents performing the corresponding navigation tasks.
The quality of the generated annotations and video-clip pairings are evaluated with (a) human evaluation studies of \textsc{NAVCON} samples, (b) training and testing of a navigation concept classifier trained on \textsc{NAVCON}, and (c) a battery of experiments with few-shot learning using GPT-4o. This work makes the following contributions:
\begin{itemize}
\item Public release of a VLN corpus containing annotations of navigation concepts and video-clip pairings for the over-$30,000$ instructions in the R2R and RxR datasets. 
\item A high-performing navigation concept classifier model for predicting navigation concepts in unseen text.
\item Extensive experiments comparing the few-shot learning of GPT-4o against the fine-tuned classifier model for predicting navigation concepts.
\end{itemize}

The paper is organized as follows. In Section \ref{background},  we discuss prior work in connection to  VLN tasks and persisting challenges which motivate our work and the need for the \textsc{NAVCON} multi-modal vision-language resource. In Section \ref{concepts}, we present the language navigation concepts annotated in \textsc{NAVCON}. 
Section \ref{annotations} describes our approach to generating silver annotations and reports a human evaluation study.  To provide further evidence of the quality, validity, and reliability of the annotations in \textsc{NAVCON}, Section \ref{models} reports on (a) training and testing a navigation concept classifier (NCC) model using \textsc{NAVCON} (b) experiments on few-shot learning with GPT-4o as a concept classifier. 
\section{Background and motivation}
\label{background}
The problem of instruction following for navigation has drawn significant attention in a wide range of domains. These include Google Street View Panoramas~\cite{chen2019touchdown}, simulated environments for quadcopters~\cite{blukis2018mapping,misra2018mapping}, multilingual settings~\cite{ku2020room}, interactive instruction following task~\cite{shridhar2020alfred,shridhar2020alfworld,min2021film,kim2021agent}, interactive vision-dialogue setups~\cite{zhu2021self}, real world scenes~\cite{anderson2021sim}, and realistic simulations of indoor scenes~\cite{anderson2018vision,krantz2020beyond,chang2017matterport3d}. One of the biggest challenges in following instructions is grounding the natural language to a sequence of observed visual inputs.

Earlier works~\cite{matuszek2010following,tellex2011understanding,kollar2010toward} assumed that the structure and semantics of the environment were available and employed topological maps and probabilistic models to associate instructions to paths. This was usually preceded by a language parsing system that would decompose the instruction to a sequence of spatial description clauses~\cite{kollar2010toward} or path description language~\cite{matuszek2010following}. With the introduction of the Vision-and-Language Navigation task~\cite{anderson2018vision}, several methods have shifted focus to solving this problem in realistic and previously unseen environments~\cite{krantz2020beyond,wang2018look,ku2020room}.

In spite of the significant progress achieved on the VLN task on public benchmarks~\cite{gao2023adaptive,an2022bevbert,chen2024mapgpt}, especially in regards to the mapping and planning~\cite{huang2023visual,krantz2022sim2sim,georgakis2022cross,an2024etpnav} aspects of the problem, the limited availability of human instruction data still remains a challenge. Recent efforts attempted to alleviate this issue by collecting large numbers of non-navigation related image-caption pairs from the web~\cite{majumdar2020improving,guhur2021airbert}. Instead of relying on task-agnostic data, other methods proposed to enrich current datasets through synthetic generation processes either by augmenting existing instructions~\cite{kamath2023new} using navigation instruction generators~\cite{wang2022less}, using semantic maps~\cite{li2024semantic}, or through a speaker model that takes as input sequences of images and semantic labels~\cite{chen2022learning}.
Another line of work ignores the collection of new data altogether and elects to take advantage of the rich pre-trained representations of large vision-language models~\cite{huang2023visual,gadre2023cows,shah2023lm,dorbala2022clip} such as CLIP~\cite{radford2021learning}. 
While all the aforementioned efforts seek to increase access to data or to aligned vision-language representations, they do not perform any analysis of the underlying navigation concepts in the instructions.

Recently, there has been a number of efforts to use additional linguistic information derived from the dependency parse of a navigation instruction to help in improving performance of the VLN agent, using specialized architectures \cite{li2021improving,zhang2023vln,zhang2022explicit,zhang2021towards}. However, the previous works employ several heuristics in addition to the noisy, off-the-shelf dependency parse of the instruction which hurts reproducibility as well as linguistic interpretability as to which parts of the instruction actually help. 

To tackle these challenges, in this work we propose a Navigation Concept Corpus (\textsc{NAVCON}) where we annotate the instructions in VLN benchmarks with cognitively motivated concept category tags, and comprehensively evaluate our annotation using human annotators to provide a reliable, general-purpose resource with annotation models that can be used for general language-guided navigation applications.
\section{Cognitively Inspired Definition of Navigation Concept Classes} \label{concepts}

We were inspired by brain mapping studies for the identification of core navigation concept classes. Navigation is a basic survival capability that the brain of animals and humans devotes dedicated areas for representing places, headings, boundaries, and path integration:

\begin{enumerate}
\item Firing rates of place cells in the hippocampus area of animals increase or decrease based on the animal’s location (Situate Yourself) \cite{o1971hippocampus}.
\item Head direction cells found in limbic system areas \cite{taube2007head} fire on the basis of the animal’s head direction independently of the animal’s location (Change Direction).
\item Boundary cells fire on the basis of boundaries of regions in the environment \cite{solstad2008representation}  (Change Region).
\item Spatial navigation and orientation rely on locomotion (and its activation of motor, vestibular, and proprioceptive systems). \cite{taube2013navigation}. From a psychological perspective, \cite{lynch1964image}'s classification of spatial behavior includes “paths” as the channels along which individual moves. (Move along a Path).
\end{enumerate}

Following these findings, we identify the four concepts above, namely "situate yourself", "change direction", "change region", and "move along a path" as core navigation concepts (shown in (Table~\ref{concept_class}). Our first task in building \textsc{NAVCON} is to identify and annotate the diverse linguistic realizations of these concepts so that we can train a model that can identify them in a variety of seen and unseen navigation instructions. We detail this effort in Section \ref{annotations}. Table \ref{concept_class} shows the four navigation concept classes included in \textsc{NAVCON}, the annotation labels, and an example of each class. 
\begin{table*}
  \centering
  \begin{tabular}{lll}
    \hline
    \textbf{Concept Class} & \textbf{Abbreviation}& \textbf{Example}\\ 
    \hline
    Situate Yourself & SIT & standing in front of that pillar\\ 
   Move along a Path & MOVE & step into this area with a large pool\\ 
     Change Direction & CD & turn around from the bench\\ 
     Change Region & CR & enter the room that is in front of you\\ 
    \hline
  \end{tabular}
  \caption{\label{concept_class}
    Navigation concept classes and example instantiations.
  }
\end{table*}
\section{Description of \textsc{NAVCON}\label{annotations}}
Research on Vision-and-Language Navigation has been based on a limited number of publicly available benchmarks. The R2R (Room-to-Room)~\cite{anderson2018vision} dataset and its R4R (Room-for-Room) and RxR (Room-Across-Rooms) \cite{ku2020room} extensions is a prominent, community-standard dataset. For this reason we opted to enrich this dataset (specifically R2R and RxR - Creative Commons Public Licenses) with annotations of the navigation concepts introduced in Section \ref{concepts}. The R2R dataset includes panoramic images of indoor spaces and written language instructions that provide guidance for navigation from one place to another within a space. The RxR dataset is multilingual with longer and more variable paths as well as more fine-grained lower level language instructions. Both R2R and RxR use simulators~\cite{savva2019habitat} built on a dataset containing real scenes~\cite{chang2017matterport3d}.
Other popular VLN datasets include Touchdown \cite{chen2019touchdown}, 
ALFRED \cite{shridhar2020alfred}, 
VLN-CE \cite{krantz2020beyond} which is R2R but ported in continuous environments rather than a navigation graph, and REVERIE \cite{qi2020reverie}. REVERIE includes short high-level instructions for executing household tasks 
where the end location is an object, while ALFRED operates in a purely synthetic environment. 



We have determined that R2R and RxR are the most suitable datasets for building \textsc{NAVCON}, due to their detailed instructions and challenging paths situated in real environments.

\begin{figure*}
    \centering
    \includegraphics[width=\textwidth]{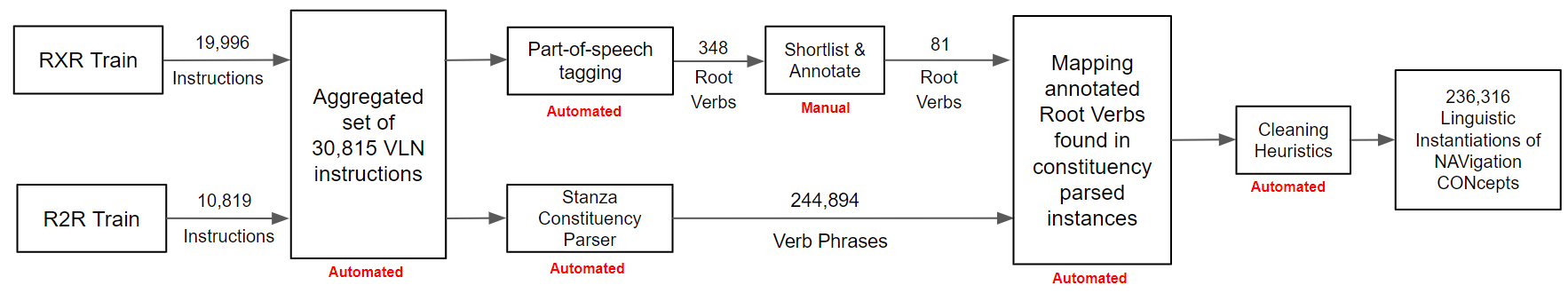}  
    \caption{Outline of processing steps for the generation of \textsc{NAVCON} annotations}
    \label{fig:img1}
\end{figure*}

\textsc{NAVCON} contains annotations of instructions taken from the following two  VLN datasets\footnote{https://github.com/jacobkrantz/VLN-CE}: a) R2R VLNCE: Room-to-Room Vision and Language Navigation in Continuous Environments \cite{krantz2020beyond} and b) RxR VLNCE: Room-Across-Room Vision and Language Navigation in Continuous Environments \cite{ku2020room}. Table \ref{table:2} shows the number of English instructions included in the two datasets across four data splits. We processed and generated annotations for the total of 30,815 instructions in the training split of the two datasets. The methodology we used to generate these large scale annotations with minimal human annotation effort is detailed in the next subsections.  Figure \ref{fig:img1} outlines all the processing steps.

\begin{table*}
  \centering
  \begin{tabular}{lllll}
    \hline
    \textbf{Dataset} & \textbf{Train} & \textbf{Test} & \textbf{Val Seen}  & \textbf{Val Unseen} \\
    \hline
    R2R VLNCE & 10,819 & 3,408 & 778 & 1,839\\ 
    RxR VLNCE & 19,996 & 3,186 & 2255 & 3,669 \\
    \hline
  \end{tabular}
  \caption{\label{table:2}
    Size of English instructions in the R2R \& RxR datasets
  }
\end{table*}

\subsection{Identification and Annotation of Navigation Concepts}
For the identification of linguistic realizations of the four navigation concept classes we defined the following subtasks: a) retrieve and evaluate unique root verbs, b) generate a list of navigation verbs, c) retrieve and clean up the verb phrases anchored by the root verbs including syntactic children. 

For the retrieval of unique root verbs we ran a full NLP pipeline including tokenization, lemmatization, part-of-speech tagging\footnote{https://spacy.io/usage/linguistic-features}, and syntactic parse  for 30,815 R2R and RxR instructions. For the syntactic parse, we used the Stanza constituency parser\footnote{https://stanfordnlp.github.io/stanza/constituency.html}. We identified 348 instances of root verbs (with a V  part-of-speech tag and being a syntactic head of the verb phrase). The list of 348 verbs was evaluated by humans to identify the verbs participating in a navigation instruction. This process resulted in a list of 81 root navigation verbs unambiguously mapped into the four navigation concepts. The distribution of the 81 ROOT VERBS in the taxonomy across the four concept classes: Situate: 29 (36\%), Move: 33 (40\%), Change Direction: 11 (14\%), and Change Region: 8 (10\%). 

Next, using the output of the syntactic parse, we retrieved all the instances of the 81 ROOT VERBS and their syntactic children. A total of 244,894 verb phrases headed by a \textsc{NAVCON} root verb were retrieved.  Note that although the range of identified root verbs is limited, the range of children which often include important navigation landmarks and other words co-occurring with \textsc{NAVCON} roots is broad. 
After a cursory analysis of the output and  clean-up (e.g., removing instances of phrases with two root verbs), we obtained automatic annotations of {\bf 236,316 linguistic instantiations of navigation concepts}.

\begin{figure*}
    \centering
    \includegraphics[width=0.9\textwidth]{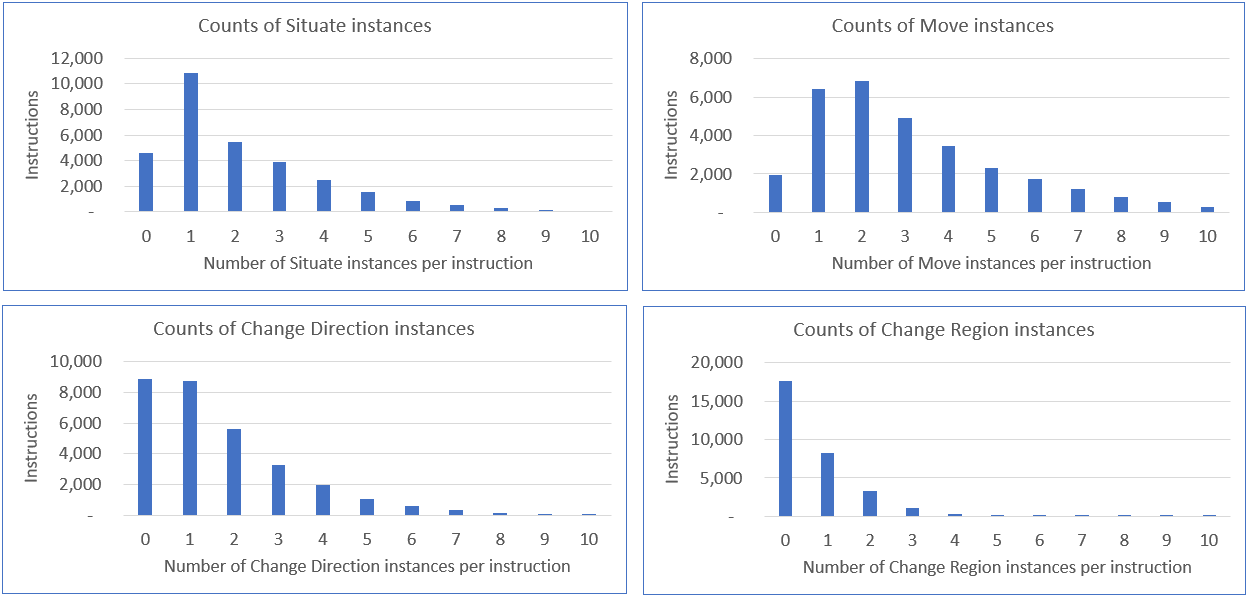}
    \caption{Distribution of navigation concepts in \textsc{NAVCON}. Instruction counts and percentages for more than 10 instances per instruction have been omitted from the charts for visual clarity.}
    \label{fig:img2}
\end{figure*}

Figure \ref{fig:img2} illustrates the distribution of the four classes of navigation concepts across the 30,815 instructions included in \textsc{NAVCON}. Instruction counts and percentages for more than 10 instances per instruction were omitted from the charts for visual clarity. The following number of instructions that have at least 1 occurrence of each concept: Situate: 10,811 (35\%), Move: 6,397 (21\%), Change Direction: 8,734 (28\%), and Change Region: 8,299 (27\%). In addition, we counted 8,560 instructions that have at least one occurrence of all four concepts. This distribution confirms that \textsc{NAVCON} includes sufficient representation of all the concept classes. We also measured the occurrences of individual root verbs in \textsc{NAVCON} instantiations. These numbers are proportionate to the reported concept distributions. A total of 70 root verbs were present in the final set out of the 81 included in our taxonomy. 
We note here the top 5 root verbs and their frequency of occurrence in \textsc{NAVCON}: 1) TURN: 51,768,  2) WALK: 29,848, 3) MOVE: 24,250, 4) GO: 22,081, and 5) FACE: 18,210. 

\begin{table}
  \centering
  \resizebox{\columnwidth}{!}{ 
    \begin{tabular}{lll}
      \hline
      \textbf{Concept} & \textbf{\makecell[l]{Annotated \\ Phrases}} & \textbf{Percentage} \\
      \hline
      Situate & 65,765 & 28\% \\ 
      Move & 98,418 & 42\% \\ 
      Change Direction & 51,798 & 22\% \\ 
      Change Region & 20,335 & 9\% \\ 
      \hline
    \end{tabular}
  }
  \caption{Distribution of \textsc{NAVCON} annotations}
  \label{table:3}
\end{table}

\subsection{Human Evaluation of Automatically Annotated Concepts}

To evaluate the quality of annotations generated by the methodology described in the previous section, we conducted a human evaluation study. We randomly extracted 300 instances of silver concept annotations resulting from our approach. Two annotators evaluated the accuracy of a) the detected text and b) the assigned concept label. The annotators were presented with a) the full text of the instruction, b) the identified concepts, and c) the text corresponding to each identified concept. Each identified concept was marked as {\em correct}, {\em incorrect}, or {\em missing}. A concept was marked missing when the instruction included a concept that was missed by our silver annotation method. Each textual span associated with a concept was evaluated as {\em correct}, {\em partially correct}, or {\em missing}. As shown in Table \ref{table:4}, our method of identifying concept instantiations missed identifying less than 4\% of instantiated concepts and correctly identified the corresponding textual spans for over 95\% of cases.


To further evaluate the quality of the textual annotations in \textsc{NAVCON}, we trained and tested a concept identification model using \textsc{NAVCON} annotations presented in Section \ref{language training}. 
\begin{table}
  \centering
  \resizebox{\columnwidth}{!}{ 
    \begin{tabular}{lll}
      \hline
      \textbf{Overlap} & \textbf{Textual Span} & \textbf{Concept Class}\\
      \hline
      Correct & 95.49\% & 95.82\% \\ 
      Incorrect & 0.31\% & 0.88\% \\ 
      Partial & 0.79\% & N/A \\ 
      Missing & 3.39\% & 3.28\% \\ 
      \hline
    \end{tabular}
  }
  \caption{Human evaluation of automatically annotated concepts and corresponding textual spans}
  \label{table:4}
\end{table}
\subsection{Identification and Annotation of Concepts on Video}\label{concept-video}
To enable training of cross-modal models for VLN tasks, \textsc{NAVCON} includes video frames corresponding to the annotated navigation concepts. To identify and retrieve the corresponding video frames, we utilized the timestamps that are included in the text instructions in the RxR dataset. The timestamp of each word corresponds to the time when that part of instruction was uttered. We utilized this mapping to annotate the correspondence of a linguistic instantiation of a navigation concept (the entire phrase) with the corresponding sequence of images. Henceforth, we will refer to a sequence of video frames corresponding to a navigation concept as a video-clip. 
 




Specifically, the video frames are extracted from the RxR dataset using the Habitat Simulator\footnote{https://aihabitat.org/}. First, we downloaded the guide annotations and their corresponding pose traces from RxR. Guide annotations contain the instruction id and scene id which allowed us to extract the scene from Matterport 3D for each episode. For each instruction, we initialized an agent in Habitat and obtained agent positions and rotations from pose traces of the instruction. At each timestep, we configured the agent with the current pose in Habitat using previously obtained position and rotation parameters. Once we placed the agent, we obtained RGBD observations through its camera and stored them as images. We repeated this process for all instructions. 

We encountered several challenges during the image retrieval process. Firstly, not all words in the RxR dataset had matching timestamps. For those that were missing, we interpolated the timestamps based on the nearest neighbors. Secondly, in some cases, phrases instead of words had timestamps mapped. We split the phrase at white-space into words so that each word received the corresponding timestamp. Thirdly, some timestamps-word mappings in RxR are not accurate which led to alignment issues when extracting the video frames. We addressed this problem by extending the timestamp window at the end of each clip to include the concept content (eg.movement) in extracted frames.  

During image retrieval, we sampled 1 out of every 10 consecutive video frames to include in the \textsc{NAVCON} dataset. Sampling the frames significantly reduced dataset size and allowed much faster rendering while retaining visual continuity. We set the output image resolution to 320 px x 240 px for retrieval speed and storage size considerations. A total of 7.6 million frames were extracted from RxR using Habitat Simulator and 2.7 million frames were mapped to corresponding concepts. As a result, we are able to include in the release of \textsc{NAVCON} 2.7 million video frames paired with corresponding concepts for 19,074 instructions. To our knowledge, \textsc{NAVCON} is the first dataset that offers paired text-video annotations of core navigation concepts at this scale.

\begin{figure*}
    \centering
    \includegraphics[width=0.85\textwidth]{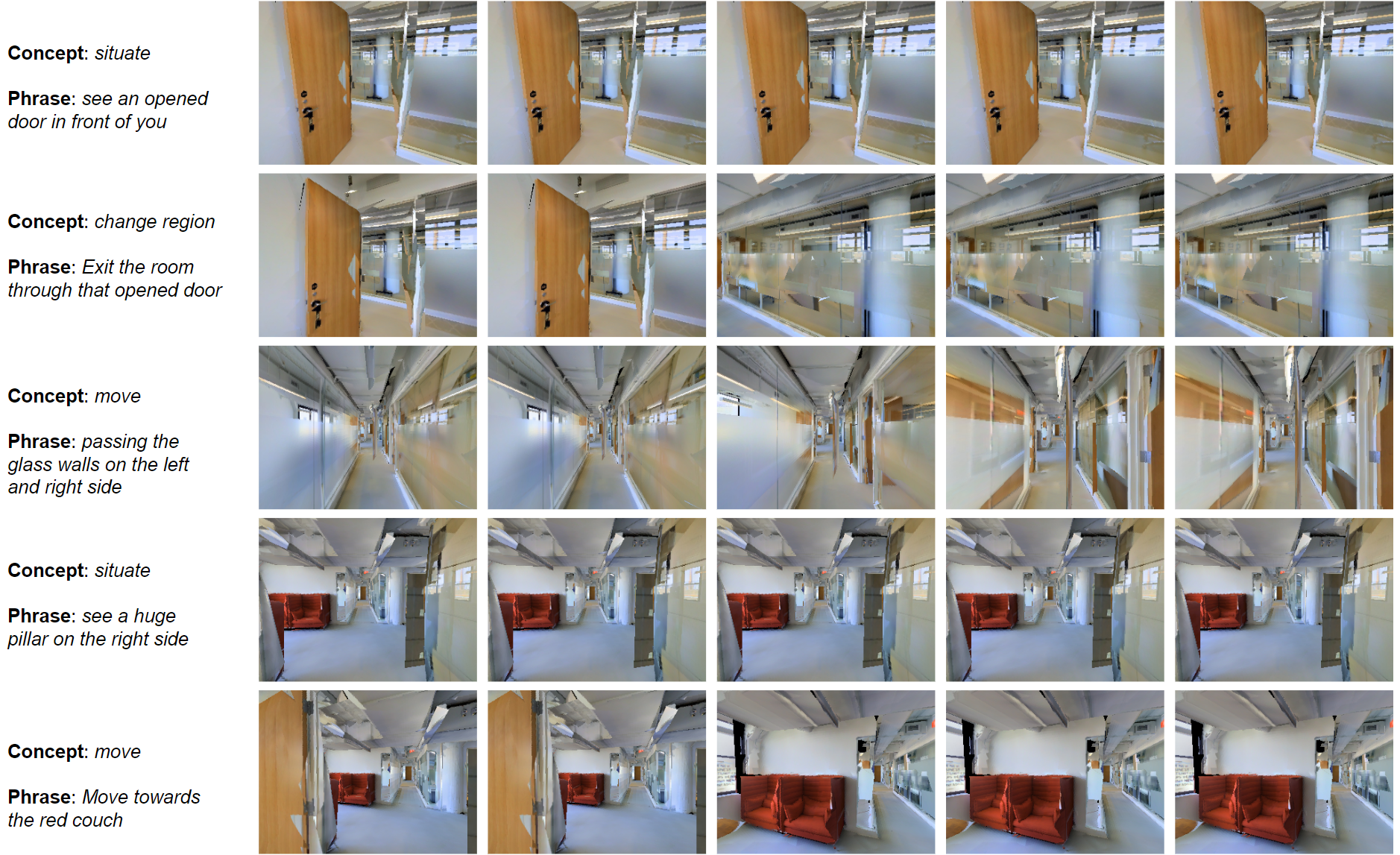}
    \caption{Example of concept-clip alignments in \textsc{NAVCON}. Timesteps progress from left to right.}
    \label{fig:clip-1}
\end{figure*}
\subsection{Evaluation of Concept-Video Pairings}

The evaluations are done by human annotators on 100 clips from 10 randomly sampled instructions. The annotators were presented with the concepts and the corresponding video frames for each concept. The annotators evaluated if the aligned sequence was an {\em exact match}  or {incomplete} match of the corresponding navigation concept. An {\em exact match} is when the sequence of images displays the navigation concept. For example, an exact match pairing of the annotation {\em you are standing in front of the mirror} would show a sequence of images displaying a mirror. An exact pairing for {\em walk towards the living room} would show a sequence of images showing that the agent is moving closer to the living-room and so forth.  We show an example of exact match instances in Figure \ref{fig:clip-1}.  An {\em incomplete match} is when the sequence of images is a partial match of the corresponding concept. For example, if for the annotation of {\em turn left into the living-room}, the sequence of images does not show the living room or any other indication of turning.

Incomplete matches occurs when the agent's pose changes at a specific time step do not align with the words corresponding to the movement as the accuracy of concept-video pairings relies heavily on the accuracy of word-timestamp mappings available in the RxR dataset. In these cases, we would need to extract extra frames in order to achieve an accurate pairing. We found that around 44\% of clips classified as movement (i.e. "move" and "change region") did not have any changes in their corresponding pose sequences. We analyzed the problematic clips and found that extending the extraction time window by 5$\%$ of the entire video sequence length increased the prediction accuracy from 73.6$\%$ to 88.6$\%$ (see Table \ref{table:clip}). This is because extending the extraction window helped with including delayed pose changes that occurred in about half of the clips with movement concepts due to inaccurate timestamp-word mapping of the RxR data that our dataset built upon.
\begin{table*}
  \centering
    \begin{tabular}{lll}
      \hline
      & \% \textbf{Clips before window extension} & \textbf{\% Clips after window extension} \\
      \hline
      Exact Match & 73.63\% & 88.62\% \\
      Incomplete Match & 26.37\% & 11.38\% \\
      \hline
    \end{tabular}
  \caption{Accuracy of concept-clip alignments}
  \label{table:clip}
\end{table*}
\section{Training Concept Navigation Models}\label{models}
\subsection{Navigation Concept Classifier (NCC)} \label{language training}
The success of a model trained on annotated datasets reflects the quality of the provided annotations. To further evaluate the quality and usefulness of the annotations in \textsc{NAVCON}, we trained a model which identifies navigation concepts and their corresponding phrases.  To achieve this, we trained a Navigation Concept Classifier (NCC) to predict for each word in the input text if it belongs to one of five classes; the four navigation concepts and one out-of-class. We fine-tuned a light-weight general purpose language representation model, distilbert-base-uncased\footnote{https://huggingface.co/distilbert-base-uncased} ~\cite{sanh2019distilbert} using 30,629 annotated instructions. 
For the training, each word was assigned a label out of the five classes mentioned above. 

\begin{figure*}[t]
  \includegraphics[width=0.48\linewidth]{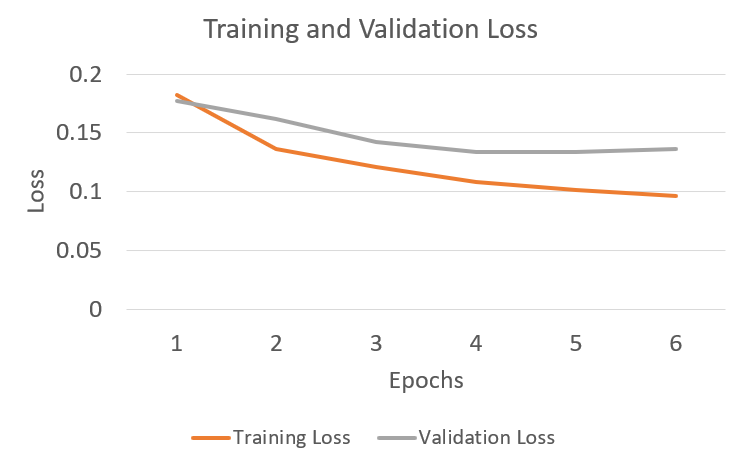} \hfill
  \includegraphics[width=0.48\linewidth]{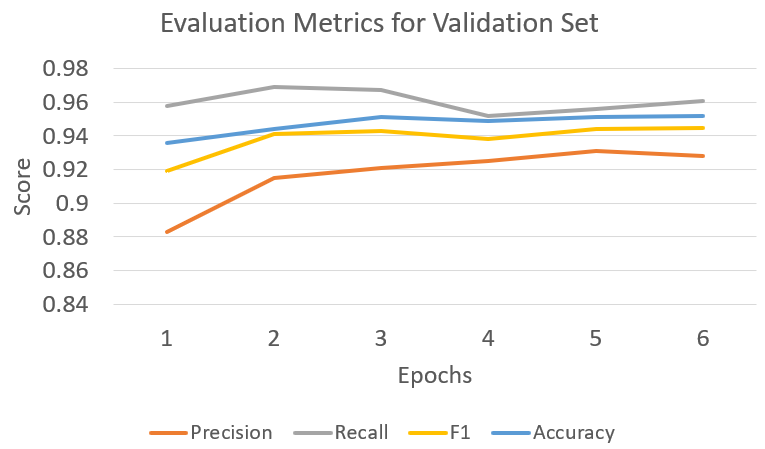}
  \caption {Navigation Concept Classifier (NCC) training and validation characteristics}
  \label{fig:training}
\end{figure*}

Distilbert-base-uncased requires a BIO  format for training data. A prefix "B-" was added for the first word of the concept phrase signifying beginning of the concept phrase and prefix "I-" was added for the remaining words in the concept phrase. We fine-tuned the classifier with a learning rate of 10\textsuperscript{-5} over 6 epochs. The training and validation set characteristics are shown in Figure \ref{fig:training}. For the cross validation evaluation, we tested both the correct identification of the phrase instantiating a navigation concept and the prediction of the concept type.  For every instruction, we first evaluated the accuracy of the predicted label. If a concept was predicted correctly, then we measured the percentage of intersection between the ground truth concept phrase and the predicted concept phrase. If a concept was not predicted correctly, then this instance and the corresponding phrase were measured as erroneous. The percentage of overlap between the ground truth phrase and the predicted phrase was measured as a ratio of overlapping tokens to all the tokens of the ground truth data, see Table \ref{table:5}. Specifically, we show the results for 100\%, 75\%, 50\%, 25\% and 0\% phrase match. The category of 0\% includes all the concepts that were not predicted correctly regardless of the corresponding phrase. In other words, we assumed 0\% phrase match for incorrect concepts.  We share the evaluation outputs with the \textsc{NAVCON} dataset. The performance of the navigation concept classifier exceeded our expectation, yielding a 96.53\% accuracy for predicting the concept and the exact phrase that corresponds to the concept.

Training a concept classifier model that successfully identifies core navigation concepts and the textual spans that describes them provides solid evidence that our approach to generating silver annotations of navigation concepts at scale is valid and that the annotations are reliable.


\subsection{Few Shot Learning with GPT-4o}\label{gptstudy}

Given the recent success of Large Language Models in few-shot learning, it is reasonable to ask whether we could use few shot learning with LLMs to generate silver annotations of high level navigation concepts. In preliminary studies that we ran with GPT-3 text-davinci-003 and GPT-4, the performance was not worthy of consideration. However, GPT-4o showed to be promising, but not as good as the NLP pipeline method. 


In what follows, we report the results of identifying navigation concepts by using few-shot learning with GPT-4o. Specifically, we use the prompt used in \ref{prompt}. The prompt describes the task of predicting navigation concepts, followed by 3 examples of navigation instructions and their corresponding navigation concepts.

\begin{table}
  \centering
  \resizebox{\columnwidth}{!}{ 
    \begin{tabular}{lll}
      \hline
       \textbf{\makecell[l]{\% Concept \\ Phrase Overlap}} & \textbf{\makecell[l]{\# \\ of Concepts}} & \textbf{\makecell[l]{\% \\ of Concepts}} \\
      \hline
      0 & 7,842 & 3.41\% \\ 
      1-25 & 3 & 0.00\% \\ 
      25-50 & 16 & 0.01\% \\ 
      50-75 & 50 & 0.02\% \\ 
      75-99 & 69 & 0.03\% \\ 
      100 & 222,043 & 96.53\% \\ 
      \hline
    \end{tabular}
  }
  \caption{Evaluation of Navigation Concept Classifier (NCC)}
  \label{table:5}
\end{table}
To evaluate the predictions from the GPT-4o model we adopted the same approach as stated in section \ref{language training}. We passed the prompt (see Appendix \ref{prompt}) along with 3 instructions from the gold dataset (refer Table \ref{table:4}). These instructions are part of \textsc{NAVCON} and were validated by the annotators. Once, GPT-4o received the tokenized and tagged few-shot examples, it learned the associations between key phrases and navigation concepts. We then instructed GPT-4o to assign concept tags to each of the words of 190 unseen instructions (from the same gold dataset) that had 1,521 concepts.
\begin{table}
  \centering
  \resizebox{\columnwidth}{!}{ 
    \begin{tabular}{lll}
      \hline
       \textbf{\makecell[l]{\% Concept \\ Phrase Overlap}} & \textbf{\makecell[l]{\# \\ of Concepts}} & \textbf{\makecell[l]{\% \\ of Concepts}} \\
      \hline
      0 & 270 & 17.75\% \\ 
      1-25 & 0 & 0.00\% \\ 
      25-50 & 0 & 0.00\% \\ 
      50-75 & 2 & 0.13\% \\ 
      75-99 & 0 & 0.00\% \\ 
      100 & 1249 & 82.12\% \\ 
      \hline
    \end{tabular}
  }
  \caption{Evaluation of Generations by GPT-4 model after few shot examples from \textsc{NAVCON}}
  \label{evaluations_gpt_2}
\end{table}



In Table \ref{evaluations_gpt_2} we see that GPT-4o performance from few-shot \textsc{NAVCON} yielding a 82.12\% accuracy for predicting the complete concept and the exact phrase.
%

\section{Limitations}
\textsc{NAVCON} primarily relies on Stanza Constituency Parser to extract verb phrases and we understand the accuracy of such parsers is not perfect. 
Furthermore, the video frames extracted are limited by the accuracy of word-timestamp mapping in the RXR dataset as well as the rendering artifacts of the Habitat simulator. 

\section{Conclusion}
We introduced four core, cognitively motivated and linguistically grounded, NAVigation CONcepts. We suspect that these four concepts may account for most if not all naturally occurring navigation instructions. We created \textsc{NAVCON}, a corpus with annotations of navigation concepts for thirty thousand navigation instructions in the R2R and RxR benchmarks. The corpus counts 200K language instantiations of navigation concepts. We paired the language navigation annotations with over 2.7 million corresponding video-clip frames. The dataset is released under the Attribution-ShareAlike 4.0 International (CC BY-SA 4.0). We provided evidence for the quality and usefulness of the annotations with a) human evaluation studies and b) experiments with few-shot learning of GPT-4o. \\
We anticipate that future VLN studies will benefit from \textsc{NAVCON} in two important ways: 1) It will render results more interpretable by examining the association of actions to higher level navigation concepts, and 2) the intermediate representation of the navigation concepts will facilitate simpler alignment of language instructions to sequential visual inputs.

\bibliography{references}








\appendix

\section{Appendix / supplemental material}
\subsubsection{Experiments with GPT-4o}\label{prompt}
Prompt for GPT-4o fewshot: 
{\em
\small
\textit{
I will give you a navigation instruction in a list format. Follow the same tokens, don't change, add, or remove anything from the list. The first element of the tuples in Values List is the tokenized words from the instruction in sequence. The second element are their acronyms: O means out of class, CR means change region, CD means change direction, SIT means situate, MOVE means move along the path. The four acronyms (CR, CD, SIT, MOVE) have a prefix in the json, which is B- or I-. B- means it is in the beginning of the phrase. I- means it is in the remaining part of the phrase. Please add the value tag to the instruction. Both the Instruction List and the Values List must have the same count. \\
Instruction: ['right', 'now', 'you', 'are', 'standing', 'at', 'the', 'entrance', 'of', 'the', 'room', 'in', 'which', 'you', 'have', 'two', 'white', 'tables', 'and', 'two', 'black', 'chairs,', 'turn', 'around', 'and', 'exit', 'the', 'room', 'through', 'the', 'open', 'door', 'and', 'now', 'turn', 'right', 'and', 'walk', 'straight', 'on', 'the', 'right', 'hand', 'side', 'you', 'can', 'see', 'the', 'open', 'door,', 'enter', 'into', 'it,', 'it', 'is', 'an', 'empty', 'room', 'turn', 'around', 'and', 'exit', 'the', 'room', 'through', 'the', 'same', 'entrance', 'door', 'and', 'now', 'turn', 'left', 'and', 'go', 'straight', 'now', 'you', 'have', 'crossed', 'the', 'semicircle', 'round', 'pillar,', 'go', 'straight', 'keep', 'going', 'straight,', 'now', 'you', 'have', 'been', 'crossed', 'the', 'second', 'semi', 'circle', 'pillar', 'keep', 'moving', 'straight', 'now', 'on', 'the', 'left', 'hand', 'side', 'you', 'can', 'see', 'the', 'semi', 'circle', 'third', 'pillar', 'and', 'now', 'turn', 'right,', 'you', 'can', 'see', 'the', 'open', 'door,', 'enter', 'into', 'the', 'open', 'door', 'room', 'and', 'move', 'towards', 'the', 'carton', 'boxes', 'which', 'are', 'present', 'under', 'the', 'white', 'table', 'right', 'in', 'front', 'of', 'the', 'carton', 'boxes', 'in', 'the', 'middle', 'of', 'the', 'room', 'that', 'is', 'the', 'end', 'point.']\\
Values List: [('right', 'O'), ('now', 'O'), ('you', 'O'), ('are', 'O'), ('standing', 'B-SIT'), ('at', 'I-SIT'), ('the', 'I-SIT'), ('entrance', 'I-SIT'), ('of', 'I-SIT'), ('the', 'I-SIT'), ('room', 'I-SIT'), ('in', 'I-SIT'), ('which', 'I-SIT'), ('you', 'I-SIT'), ('have', 'I-SIT'), ('two', 'I-SIT'), ('white', 'I-SIT'), ('tables', 'I-SIT'), ('and', 'I-SIT'), ('two', 'I-SIT'), ('black', 'I-SIT'), ('chairs,', 'I-SIT'), ('turn', 'B-CD'), ('around', 'I-CD'), ('and', 'O'), ('exit', 'B-CR'), ('the', 'I-CR'), ('room', 'I-CR'), ('through', 'I-CR'), ('the', 'I-CR'), ('open', 'I-CR'), ('door', 'I-CR'), ('and', 'O'), ('now', 'O'), ('turn', 'B-CD'), ('right', 'I-CD'), ('and', 'O'), ('walk', 'O'), ('straight', 'O'), ('on', 'O'), ('the', 'O'), ('right', 'O'), ('hand', 'O'), ('side', 'O'), ('you', 'O'), ('can', 'O'), ('see', 'B-SIT'), ('the', 'I-SIT'), ('open', 'I-SIT'), ('door,', 'I-SIT'), ('enter', 'B-CR'), ('into', 'I-CR'), ('it,', 'I-CR'), ('it', 'O'), ('is', 'O'), ('an', 'O'), ('empty', 'O'), ('room', 'O'), ('turn', 'B-CD'), ('around', 'I-CD'), ('and', 'O'), ('exit', 'B-CR'), ('the', 'I-CR'), ('room', 'I-CR'), ('through', 'I-CR'), ('the', 'I-CR'), ('same', 'I-CR'), ('entrance', 'I-CR'), ('door', 'I-CR'), ('and', 'O'), ('now', 'O'), ('turn', 'B-CD'), ('left', 'I-CD'), ('and', 'O'), ('go', 'O'), ('straight', 'O'), ('now', 'O'), ('you', 'O'), ('have', 'O'), ('crossed', 'O'), ('the', 'O'), ('semicircle', 'O'), ('round', 'O'), ('pillar,', 'O'), ('go', 'B-MOVE'), ('straight', 'I-MOVE'), ('keep', 'I-MOVE'), ('going', 'I-MOVE'), ('straight,', 'I-MOVE'), ('now', 'O'), ('you', 'O'), ('have', 'O'), ('been', 'O'), ('crossed', 'O'), ('the', 'O'), ('second', 'O'), ('semi', 'O'), ('circle', 'O'), ('pillar', 'O'), ('keep', 'B-MOVE'), ('moving', 'I-MOVE'), ('straight', 'I-MOVE'), ('now', 'O'), ('on', 'O'), ('the', 'O'), ('left', 'O'), ('hand', 'O'), ('side', 'O'), ('you', 'O'), ('can', 'O'), ('see', 'O'), ('the', 'O'), ('semi', 'O'), ('circle', 'O'), ('third', 'O'), ('pillar', 'O'), ('and', 'O'), ('now', 'O'), ('turn', 'B-CD'), ('right,', 'I-CD'), ('you', 'O'), ('can', 'O'), ('see', 'B-SIT'), ('the', 'I-SIT'), ('open', 'I-SIT'), ('door,', 'I-SIT'), ('enter', 'B-CR'), ('into', 'I-CR'), ('the', 'I-CR'), ('open', 'I-CR'), ('door', 'I-CR'), ('room', 'I-CR'), ('and', 'O'), ('move', 'B-MOVE'), ('towards', 'I-MOVE'), ('the', 'I-MOVE'), ('carton', 'I-MOVE'), ('boxes', 'I-MOVE'), ('which', 'I-MOVE'), ('are', 'I-MOVE'), ('present', 'I-MOVE'), ('under', 'I-MOVE'), ('the', 'I-MOVE'), ('white', 'I-MOVE'), ('table', 'I-MOVE'), ('right', 'I-MOVE'), ('in', 'I-MOVE'), ('front', 'I-MOVE'), ('of', 'I-MOVE'), ('the', 'I-MOVE'), ('carton', 'I-MOVE'), ('boxes', 'I-MOVE'), ('in', 'I-MOVE'), ('the', 'I-MOVE'), ('middle', 'I-MOVE'), ('of', 'I-MOVE'), ('the', 'I-MOVE'), ('room', 'I-MOVE'), ('that', 'I-MOVE'), ('is', 'I-MOVE'), ('the', 'I-MOVE'), ('end', 'I-MOVE'), ('point.', 'I-MOVE')]\\
Instruction: ['begin', 'in', 'a', 'bedroom', 'with', 'a', 'kitchen.', 'head', 'to', 'the', 'left', 'of', 'the', 'blue', 'rug', 'next', 'to', 'the', 'fridge.']\\
Values List: [('begin', 'O'), ('in', 'O'), ('a', 'O'), ('bedroom', 'O'), ('with', 'O'), ('a', 'O'), ('kitchen.', 'O'), ('head', 'B-MOVE'), ('to', 'I-MOVE'), ('the', 'I-MOVE'), ('left', 'I-MOVE'), ('of', 'I-MOVE'), ('the', 'I-MOVE'), ('blue', 'I-MOVE'), ('rug', 'I-MOVE'), ('next', 'I-MOVE'), ('to', 'I-MOVE'), ('the', 'I-MOVE'), ('fridge.', 'I-MOVE')]\\
Instruction: ['you', 'are', 'facing', 'a', 'wall', 'with', 'a', 'view', 'on', 'the', 'left', 'outside.', 'turn', 'right', 'and', 'walk', 'inside', 'the', 'house', 'to', 'the', 'front', 'of', 'the', 'silver', 'bar', 'stools.', 'walk', 'all', 'the', 'way', 'to', 'the', 'last', 'bar', 'stool', 'that', 'is', 'closer', 'to', 'the', 'stairs', 'on', 'the', 'right.', 'turn', 'right.', 'walk', 'between', 'the', 'dining', 'room', 'table', 'and', 'the', 'back', 'of', 'the', 'white', 'couch.', 'walk', 'around', 'the', 'white', 'couch', 'to', 'the', 'right', 'and', 'through', 'the', 'opening', 'to', 'the', 'right.', 'you', 'will', 'see', 'a', 'black', 'and', 'white', 'picture.', 'turn', 'left', 'and', 'take', 'the', 'first', 'right.', 'you', 'will', 'see', 'a', 'washer', 'and', 'dryer', 'in', 'a', 'room.', 'stay', 'in', 'the', 'threshold', 'of', 'the', 'door.', 'you', 'are', 'done.']\\
Values List: [('you', 'O'), ('are', 'O'), ('facing', 'O'), ('a', 'O'), ('wall', 'O'), ('with', 'O'), ('a', 'O'), ('view', 'O'), ('on', 'O'), ('the', 'O'), ('left', 'O'), ('outside.', 'O'), ('turn', 'B-CD'), ('right', 'I-CD'), ('and', 'O'), ('walk', 'B-MOVE'), ('inside', 'I-MOVE'), ('the', 'I-MOVE'), ('house', 'I-MOVE'), ('to', 'I-MOVE'), ('the', 'I-MOVE'), ('front', 'I-MOVE'), ('of', 'I-MOVE'), ('the', 'I-MOVE'), ('silver', 'I-MOVE'), ('bar', 'I-MOVE'), ('stools.', 'I-MOVE'), ('walk', 'B-MOVE'), ('all', 'I-MOVE'), ('the', 'I-MOVE'), ('way', 'I-MOVE'), ('to', 'I-MOVE'), ('the', 'I-MOVE'), ('last', 'I-MOVE'), ('bar', 'I-MOVE'), ('stool', 'I-MOVE'), ('that', 'I-MOVE'), ('is', 'I-MOVE'), ('closer', 'I-MOVE'), ('to', 'I-MOVE'), ('the', 'I-MOVE'), ('stairs', 'I-MOVE'), ('on', 'I-MOVE'), ('the', 'I-MOVE'), ('right.', 'I-MOVE'), ('turn', 'B-CD'), ('right.', 'I-CD'), ('walk', 'O'), ('between', 'O'), ('the', 'O'), ('dining', 'O'), ('room', 'O'), ('table', 'O'), ('and', 'O'), ('the', 'O'), ('back', 'O'), ('of', 'O'), ('the', 'O'), ('white', 'O'), ('couch.', 'O'), ('walk', 'B-MOVE'), ('around', 'I-MOVE'), ('the', 'I-MOVE'), ('white', 'I-MOVE'), ('couch', 'I-MOVE'), ('to', 'I-MOVE'), ('the', 'I-MOVE'), ('right', 'I-MOVE'), ('and', 'I-MOVE'), ('through', 'I-MOVE'), ('the', 'I-MOVE'), ('opening', 'I-MOVE'), ('to', 'I-MOVE'), ('the', 'I-MOVE'), ('right.', 'I-MOVE'), ('you', 'O'), ('will', 'O'), ('see', 'B-SIT'), ('a', 'I-SIT'), ('black', 'I-SIT'), ('and', 'I-SIT'), ('white', 'I-SIT'), ('picture.', 'I-SIT'), ('turn', 'B-CD'), ('left', 'I-CD'), ('and', 'O'), ('take', 'O'), ('the', 'O'), ('first', 'O'), ('right.', 'O'), ('you', 'O'), ('will', 'O'), ('see', 'B-SIT'), ('a', 'I-SIT'), ('washer', 'I-SIT'), ('and', 'I-SIT'), ('dryer', 'I-SIT'), ('in', 'I-SIT'), ('a', 'I-SIT'), ('room.', 'I-SIT'), ('stay', 'B-SIT'), ('in', 'I-SIT'), ('the', 'I-SIT'), ('threshold', 'I-SIT'), ('of', 'I-SIT'), ('the', 'I-SIT'), ('door.', 'I-SIT'), ('you', 'O'), ('are', 'O'), ('done.', 'O')]
}}

\end{document}